\documentclass[11pt]{article}
\usepackage{acl}
\usepackage{times}
\usepackage{latexsym}
\usepackage[T1]{fontenc}
\usepackage[utf8]{inputenc}
\usepackage{microtype}
\usepackage{inconsolata}
\usepackage{graphicx}
\usepackage{amsmath}
\usepackage{amssymb}
\usepackage{booktabs}

\newcommand{\C}{\textit{C}}
\newcommand{\A}{\textit{A}}
\newcommand{\Mlab}{\textit{M}}

\title{When LLMs Agree, Are They Right?\\
Auditing Self-Consistency and Cross-Model Agreement as Confidence Signals}
\author{Kaihua Ding\thanks{The author's prior work spans output-based error
  estimation \citep{ding2018dissertation} and AI-system evaluation and
  assessment design \citep{ding2025airesilient}.} \\
  University of Pennsylvania \\
  \texttt{dkaihua@upenn.edu}}

\begin{document}
\maketitle

\begin{abstract}
LLM-as-judge \citep{zheng2023judging} is increasingly the default for evaluating AI systems in
enterprise pipelines, often scaled to ensembles \citep{verga2024poll} or `mixture-of-experts'
\citep{shazeer2017moe} panels of judges. These systems share a key assumption:
that \emph{consistency}---agreement among judges, or among a model's own samples---indicates
correctness. We show this assumption is unreliable. Agreement is not accuracy: a model can agree with
itself, and different models can agree with each other, out of shared bias, a memorized heuristic, or
an option-position prior rather than truth. We ask \emph{when} agreement is nonetheless a usable proxy, in a
large-scale cross-runner study: $53$ runners drew $K{=}50$ samples for assigned overlapping cases
across comparisons of model tier, prompting, and scale on GPQA Diamond \citep{rein2024gpqa} and AIME \citep{aime}---$265{,}000$ samples.
Using majority-correctness as the deployment label and a hierarchical runner-clustered bootstrap,
agreement is a positive but weak predictor (Spearman rank correlation $\rho$ $0.20$--$0.59$, all positive under item-clustered
resampling) whose usefulness is \emph{regime-dependent}: best for unsaturated mid-tier models and for
allocating compute, and worst---over-confident yet no more accurate---for the most consistent frontier
model (agreement---the share of its $K$ samples that concur on the majority answer, i.e.\ self-consistency $\C$---$\geq0.8$ on $77\%$ of GPQA case-result entries, $48\%$ of those wrong). An exploratory
cross-family check on \texttt{claude-haiku-4.5}, \texttt{claude-sonnet-4.6}, and \texttt{claude-opus-4.7} shows the same frontier over-confidence, with confident errors
recurring across providers above a marginal-preserving null. Self-consistency is thus a \emph{conditional}
proxy for correctness, not a standalone confidence score. We publicly release the de-identified
per-run rows and answer distributions.
\end{abstract}

\section{Introduction}
\label{sec:intro}
Large language models are increasingly used to \emph{evaluate} AI systems---as
LLM-as-judge, and in ensemble or multi-judge (`mixture-of-experts') panels now common in
enterprise evaluation pipelines \citep{zheng2023judging}. These pipelines rest on a
pervasive intuition: that \emph{agreement} signals correctness---if several judges concur,
or if a model returns the same answer across many stochastic samples, the answer is trusted,
and if they disagree it is treated as a guess. But agreement-with-itself is not the same as being
right: a model can repeat an answer because it is genuinely certain, or because every sample passes
through the same memorized heuristic, shared misconception, option-position prior, or systematic
hallucination. This is the self-referential analogue of the \emph{self-preference} bias documented when
models judge their own outputs \citep{zheng2023judging,panickssery2024llmevaluators}---agreement with
oneself can encode shared bias rather than correctness. Self-consistency decoding operationalizes
this by returning the majority answer over $K$ samples \citep{wang2023selfconsistency},
building on chain-of-thought prompting \citep{wei2022cot}. The same agreement signal
is now reused well beyond decoding---as a confidence estimate for cost-based routing
and cascades \citep{chen2023frugalgpt}, for adaptive sample-budget allocation
\citep{aggarwal2023adaptive}, and for selective prediction and abstention
\citep{geifman2017selective,kamath2020selective}.

That self-consistency raises \emph{accuracy} via majority voting is well established,
as is the broader finding that elicited LLM confidence is often miscalibrated and can
even worsen after post-training \citep{kadavath2022know,lin2022teaching,tian2023justask,openai2023gpt4}.
Less systematically examined is how reliably the \emph{agreement} signal itself functions as the
confidence proxy that routing and abstention systems already assume it to be---and \emph{when} it
fails---under a controlled, cross-replicated audit on hard
reasoning benchmarks. We provide that audit, using a cross-runner design to
separate confident errors that recur across different runners and prompts from those that are sampling noise.
Our hypothesis is that \textbf{self-consistency is not accuracy, but a regime-dependent empirical proxy for
it}: a positive yet weak signal whose usefulness depends on measurable conditions---model tier and
agreement regime, answer space, and intended use---which we map. It is usable where agreement is
unsaturated (mid-tier models) and for allocating compute, but unreliable as a standalone confidence score,
worst for the over-confident frontier model.

\paragraph{Contributions.}
(1) A large-scale ($265{,}000$-sample), cross-runner audit of self-consistency
\emph{as a confidence/abstention/routing signal} on GPQA Diamond \citep{rein2024gpqa} and AIME \citep{aime}, with a hierarchical
runner-clustered bootstrap and multiplicity control.
(2) Evidence that the agreement signal \emph{degrades} on the most consistent frontier
model---over-confident but no more accurate---consistent with the known post-RLHF calibration
regression, making confidence-routing toward it counterproductive.
(3) A recurrence-based separation of recurring from stochastic confident-wrongness, plus an
option-shuffle control showing part of GPQA ``confidence'' is positional.
(4) Honest null/marginal results: chain-of-thought improves accuracy but only marginally improves
the agreement--correctness signal, and a confidence-routed cascade is dominated by always using
the mid-tier model.
(5) A released de-identified, schema-validated dataset and analysis pipeline.
(6) An \emph{exploratory}, agent-mediated cross-family check on three Claude tiers, consistent with the
frontier over-confidence and suggesting confident errors are partly \emph{shared across providers} (GPT
and Claude pick the same wrong answers above a marginal-preserving null for the two smaller tiers)---so
high agreement can reflect shared bias, not correctness.

\begin{table*}[t]
\centering\small
\setlength{\tabcolsep}{6pt}
\begin{tabular}{@{}llcccc@{}}
\toprule
Model (axis) & Benchmark & $\overline{\C}$ & $\overline{\Mlab}$ & $\rho(\C,\A)$ [95\% CI] & $\rho(\C,\Mlab)$ [95\% CI] \\
\midrule
gpt-4.1-nano (A)        & GPQA & .76 & .40 & .12\,[$-$.04,\,.29]\ \textsuperscript{$\dagger$} & .21\,[.07,\,.34] \\
gpt-4.1-nano (A)        & AIME & .39 & .23 & .42\,[.14,\,.67] & .47\,[.25,\,.64] \\
gpt-4.1-mini (A)        & GPQA & .83 & .51 & .37\,[.21,\,.53] & .30\,[.17,\,.43] \\
gpt-4.1-mini (A)        & AIME & .51 & .32 & .58\,[.28,\,.81] & .59\,[.37,\,.75] \\
gpt-4.1-mini, ZS (B)    & GPQA & .80 & .52 & .34\,[.14,\,.54] & .26\,[.11,\,.39] \\
gpt-4.1-mini, ZS (B)    & AIME & .58 & .36 & .56\,[.31,\,.80] & .55\,[.35,\,.71] \\
gpt-4.1-mini, CoT (B)   & GPQA & .79 & .58 & .54\,[.35,\,.71] & .38\,[.24,\,.51] \\
gpt-4.1-mini, CoT (B)   & AIME & .61 & .44 & .63\,[.35,\,.86] & .58\,[.37,\,.75] \\
gpt-4.1-mini (C)        & GPQA & .82 & .52 & .42\,[.25,\,.58] & .35\,[.24,\,.46] \\
gpt-4.1-mini (C)        & AIME & .46 & .24 & .48\,[.17,\,.72] & .52\,[.31,\,.69] \\
\textbf{gpt-4.1 (C)}    & GPQA & \textbf{.89} & .48 & .19\,[.06,\,.34] & .20\,[.07,\,.33] \\
\textbf{gpt-4.1 (C)}    & AIME & .46 & .17 & .27\,[.10,\,.44] & .31\,[.13,\,.45] \\
\bottomrule
\end{tabular}
\caption{Spearman rank correlation $\rho$ of self-consistency $\C$ with sample accuracy $\A$ and with
majority-correctness $\Mlab$, per cell, with hierarchical runner-clustered-bootstrap 95\% CIs
($B{=}2000$). $\overline{\C}$, $\overline{\Mlab}$ are cell means. All twelve $\rho(\C,\Mlab)$ are
positive and survive Holm correction for multiple comparisons (a family-wise adjustment across the twelve tests that guards against false positives; adj.\ $p\leq.0024$). \textsuperscript{$\dagger$}Only the
nano--GPQA $\rho(\C,\A)$ CI crosses zero---its $95\%$ interval includes $0$, so that single correlation is not statistically distinguishable from no association. The frontier gpt-4.1 has the highest agreement
($\overline{\C}{=}.89$) yet the lowest $\rho$ and no accuracy advantage over mini.}
\label{tab:rho}
\end{table*}

\begin{table}[t]
\centering\footnotesize
\setlength{\tabcolsep}{3pt}
\begin{tabular}{@{}llcc@{}}
\toprule
Post-hoc focal contrast (on $\Mlab$) & Bench & Estimate [95\% CI] & $p$ \\
\midrule
E1: Axis-B CoT, $\Delta\Mlab$ (paired) & GPQA & $+.066$\,[$-.005$,\,.132] & .022 \\
E2: Axis-C frontier, $\Delta\rho(\C,\Mlab)$ & GPQA & $-.18$\,[$-.31$,\,$-.03$] & .025 \\
\midrule
\multicolumn{4}{@{}l}{\emph{Secondary (AIME):}}\\
\ \ Axis-C frontier, $\Delta\rho(\C,\Mlab)$ & AIME & $-.26$\,[$-.40$,\,$-.12$] & .0006 \\
\ \ Axis-B CoT, $\Delta\Mlab$ (paired) & AIME & $+.080$\,[$-.009$,\,.193] & .003 \\
\bottomrule
\end{tabular}
\caption{\emph{Post hoc} focal contrasts (both GPQA) on the deployment label $\Mlab$ (E1: Wilcoxon signed-rank test on
paired $\Delta\Mlab$; E2: permutation on $\Delta\rho$); AIME secondary. These contrasts were chosen
\emph{after} earlier analyses and prior review, so the $p$-values are \textbf{descriptive}, not
confirmatory; Holm is applied only to the twelve per-cell $\rho(\C,\Mlab)$ tests of Table~\ref{tab:rho}.
E2 is the stronger contrast---it survives a Bonferroni-for-two multiple-testing correction ($p_{\text{adj}}{=}.05$) and a
case-clustered bootstrap, and recurs on AIME ($\Delta\rho{=}-.26$, $p{=}6{\times}10^{-4}$); E1 is
\emph{borderline} (its $\Delta\Mlab$ CI crosses zero), though its accuracy effect is strong
($\Delta\A{=}+.067$, $p{=}2{\times}10^{-5}$).}
\label{tab:confirm}
\end{table}

\section{Setup}
\label{sec:setup}
\paragraph{Data provenance.} The runs were produced by $53$ runners completing
an assignment in a graduate-level computer science cohort (participants shared a course context and may have exchanged code or
discussion, so runs are independently submitted but not guaranteed statistically independent); we
perform a \emph{secondary, de-identified} meta-analysis
of their submitted run-level outputs. No runner is identifiable: all personal identifiers are
removed and replaced with anonymous indices. We release de-identified per-runner--per-case rows
(an anonymous runner index, case id, the answer-count distribution, and self-consistency $\C$, sample accuracy $\A$, and the majority-correct indicator $\Mlab$---all defined under \emph{Metrics} below) \emph{without}
GPQA question text---sufficient to reproduce the clustered analysis. Because
runners were assigned overlapping cases (and each runner ran both of its axis's conditions on the same
cases, so Axis-B and Axis-C contrasts are paired within runner), each (axis, condition, case) cell is
independently re-sampled by a mean of $2.5$ runners (up to $9$); in total the data span $5{,}300$ case-result rows
over $394$ unique cases. This overlap is the basis for the replication analysis in
\S\ref{sec:intrinsic}.

\paragraph{Benchmarks.} \textbf{GPQA Diamond} \citep{rein2024gpqa}: graduate-level four-option
multiple choice ($\{A,B,C,D\}$). \textbf{AIME} \citep{aime}: integer-answer competition mathematics
($\{0,\dots,999\}$). They bracket opposite ends of an answer-space spectrum: GPQA's four options
floor $\C$ near $1/4$ even under random sampling, so high $\C$ is the discriminating regime;
AIME's large space floors $\C$ for any model that cannot reason.

\paragraph{Design.} Each runner was assigned one of three controlled comparisons, each with two
conditions ($a$ vs.\ $b$), at \texttt{temperature}$=1.0$, $K{=}50$, on the gpt-4.1 family:
\textbf{Axis A} model tier (nano vs.\ mini, zero-shot); \textbf{Axis B} prompt strategy (mini
zero-shot vs.\ mini chain-of-thought); \textbf{Axis C} mid vs.\ frontier (mini vs.\ gpt-4.1,
zero-shot). Runners implemented their own prompts, so prompt wording varies---a source of variance
we treat as part of the measurement (\S\ref{sec:limitations}). Each runner ran \emph{both} conditions
of its axis on the \emph{same} cases, so all Axis-B and Axis-C contrasts are paired within runner
($100\%$ pairing); Table~\ref{tab:design} reports the counts. Exact model snapshots and run timestamps
were not logged and are unrecoverable, so our replication is across runners and prompts, not across time
(\S\ref{sec:limitations}).

\begin{table}[t]
\centering\footnotesize
\setlength{\tabcolsep}{4pt}
\begin{tabular}{@{}llcccc@{}}
\toprule
Axis ($a$/$b$) & Bench & runners & cases & rows & rep \\
\midrule
A nano/mini       & GPQA & 18 & 178 & 900 & 2.5 \\
                  & AIME & 18 & 178 & 900 & 2.5 \\
B ZS/CoT (mini)   & GPQA & 17 & 180 & 850 & 2.4 \\
                  & AIME & 17 & 172 & 850 & 2.5 \\
C mini/gpt-4.1    & GPQA & 18 & 182 & 900 & 2.5 \\
                  & AIME & 18 & 180 & 900 & 2.5 \\
\bottomrule
\end{tabular}
\caption{Design and counts per axis. ``rows'' are case-result rows over both conditions; ``rep'' is the
mean number of runners per (condition, case) cell (max $6$--$9$). Each runner ran both conditions on the
same cases, so Axis-B/C contrasts are paired within runner. Totals: $53$ runners, $394$ unique cases,
$5{,}300$ rows.}
\label{tab:design}
\end{table}

\paragraph{Metrics.} For each (case, condition) we compute self-consistency
$\C=n_{\text{maj}}/K$, sample accuracy $\A=n_{\text{correct}}/K$, and the majority-correct
indicator $\Mlab=\mathbb{1}[\text{majority answer}=\text{ground truth}]$---the indicator that equals $1$ when the majority vote matches the ground-truth answer and $0$ otherwise. Because deployed
majority-vote systems return the majority answer, $\Mlab$ is our \textbf{primary} outcome; $\A$ is
secondary. We treat $\C$ as the deployed confidence \emph{score under audit}, not as a calibrated
probability: $\C$ has benchmark-dependent floors (near $1/4$ for four-option GPQA, far lower for AIME),
so ``over-confident'' means high $\C$ co-occurring with errors, not a miscalibrated probability in the
formal sense; expected calibration error (ECE) and Brier score of $\C$ against $\Mlab$ are operational diagnostics of that deployment use
(reliability diagrams in the appendix). \emph{Unit of analysis:} cell means, ECE/Brier, and
err$\mid\C{\geq}.8$ are \textbf{row-weighted} over case-result entries; ``always-CW'' and the
recurrence rates of \S\ref{sec:intrinsic} are \textbf{case-weighted} over unique cases.

\paragraph{Inference.} We report the Spearman rank correlation $\rho$. Entries are not i.i.d.\ (each runner contributes
many rows; each case is seen by many runners), so all CIs use a \textbf{hierarchical runner-clustered
bootstrap} (resample runners, then cases within runner; $B{=}2000$; Appendix~\ref{app:boot}). Because
this corrects runner dependence but not the globally shared-case factor, we also report a
\textbf{case-clustered} bootstrap and a leave-one-runner check for the headline frontier result
(\S\ref{sec:frontier}). Before the revised analysis we designated \textbf{two endpoints} (both on GPQA, declared
\emph{post hoc} for this revision, not preregistered)---the paired Axis-B chain-of-thought effect
(E1; the Wilcoxon signed-rank test---a paired significance test on the within-runner differences, not a correlation---on $\Delta\Mlab$) and the Axis-C frontier degradation (E2; a permutation test on
$\Delta\rho(\C,\Mlab)$); their $p$-values are reported raw, with a Bonferroni-for-two correction (a multiple-comparison adjustment for testing two endpoints) in
Table~\ref{tab:confirm}. The remaining cells are exploratory; these endpoints were declared for the revised
analysis (prompted by prior review) and are \emph{not} a preregistration. Holm correction is applied
only to the family of twelve per-cell $\rho(\C,\Mlab)$ tests (Table~\ref{tab:rho}); correlation
\emph{differences} use permutation tests. No inferential claim in this paper is preregistered; the
endpoint tests in Table~\ref{tab:confirm} are descriptive robustness summaries, not confirmatory.

\section{Does self-consistency predict correctness?}
\label{sec:headline}
Table~\ref{tab:rho} reports $\rho(\C,\cdot)$ for all twelve cells. The picture is consistent with a
\emph{conditional proxy}: the Spearman correlation between self-consistency $\C$ and majority-correctness
$\Mlab$ (whether the majority vote matches ground truth) is positive in every cell, but weak, and its
strength is \emph{conditional on} the regime---model tier, benchmark, and agreement level---that we vary next.
\textbf{(1) Positive but weak.} No cell exceeds $\rho(\C,\Mlab)=0.6$; most variance in correctness
is unexplained by agreement, so a ``high-$\C$ auto-accept'' rule still admits many wrong answers.
\textbf{(2) The signal pivots on the label.} For nano on GPQA, $\rho(\C,\A)=0.12$ has a CI crossing
zero whereas $\rho(\C,\Mlab)=0.21$ is positive---reinforcing $\Mlab$ as the right target. All twelve
$\rho(\C,\Mlab)$ are positive and survive Holm correction (adj.\ $p\leq0.0024$); they also all remain
positive and exclude zero under a \emph{case}-clustered bootstrap that resamples items rather than
runners (Table~\ref{tab:caseclust}), so the link is not an artifact of the clustering choice.
\textbf{(3) Non-monotonic in scale.} The frontier gpt-4.1 has the \emph{lowest} GPQA
$\rho(\C,\Mlab)$ ($0.20$) despite the highest mean $\C$ ($0.89$); we examine this next.

\begin{table}[t]
\centering\footnotesize
\setlength{\tabcolsep}{4pt}
\begin{tabular}{@{}lcc@{}}
\toprule
Cell (axis/cond) & GPQA $\rho(\C,\Mlab)$\,[CI] & AIME $\rho(\C,\Mlab)$\,[CI] \\
\midrule
A/$a$ nano     & .21\,[.10,.30] & .47\,[.38,.54] \\
A/$b$ mini     & .30\,[.19,.41] & .59\,[.53,.65] \\
B/$a$ mini ZS  & .26\,[.15,.35] & .55\,[.47,.62] \\
B/$b$ mini CoT & .38\,[.26,.48] & .58\,[.50,.65] \\
C/$a$ mini     & .35\,[.24,.45] & .52\,[.45,.58] \\
C/$b$ gpt-4.1  & .20\,[.09,.32] & .31\,[.21,.41] \\
\bottomrule
\end{tabular}
\caption{Item-clustered robustness: \emph{case}-clustered bootstrap 95\% CIs for all twelve
$\rho(\C,\Mlab)$ (resampling cases, not runners; $B{=}1000$). All twelve are positive and exclude zero,
so the agreement--correctness link is not an artifact of the runner-clustered choice.}
\label{tab:caseclust}
\end{table}

\section{The most consistent model is the worst-calibrated voter (in this audit)}
\label{sec:frontier}
\begin{figure}[t]
\centering
\includegraphics[width=0.82\columnwidth]{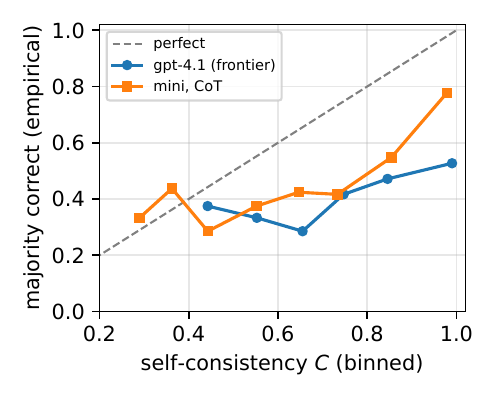}
\caption{GPQA reliability: binned self-consistency $\C$ vs.\ empirical majority-correctness. The dashed
diagonal marks perfect calibration, where binned self-consistency equals the empirical fraction correct;
points below it are over-confident. The frontier
gpt-4.1 sits well \emph{below} the diagonal at high $\C$ (majority correct only $52\%$ when $\C\geq0.8$),
i.e.\ over-confident; CoT-mini ($72\%$) tracks the diagonal more closely.}
\label{fig:reliability}
\end{figure}

\begin{figure}[t]
\centering
\includegraphics[width=\columnwidth]{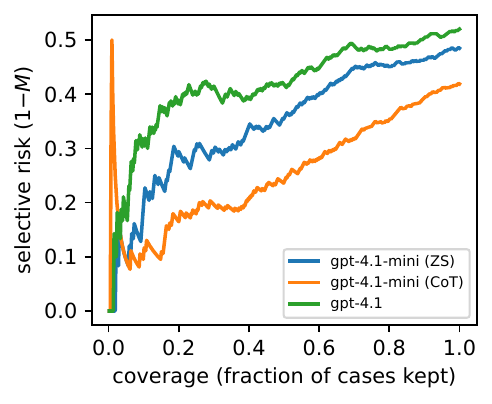}
\caption{GPQA risk--coverage by self-consistency for the three key cells. Cases are answered in order of
decreasing self-consistency $\C$; the horizontal axis is \emph{coverage} (the fraction of cases kept, most
consistent first) and the vertical axis is \emph{selective risk} (the majority-error rate, $1{-}M$, among the
kept cases)---so a good confidence signal gives a low, monotonically rising curve. The frontier gpt-4.1 curve is
highest (worst) and CoT-mini lowest, despite gpt-4.1's higher mean agreement---self-consistency orders
correctness least well exactly where agreement is highest.}
\label{fig:core}
\end{figure}

In this audit (one OpenAI model family, with an exploratory Claude check in \S\ref{sec:crossfamily}), on
the same Axis-C cases gpt-4.1 has a significantly lower agreement--correctness correlation than
mini: $\Delta\rho(\C,\Mlab)=-0.18\,[-0.31,-0.03]$ on GPQA ($p{=}0.025$, permutation) and
$-0.26\,[-0.40,-0.12]$ on AIME ($p{=}0.0006$). This is not because gpt-4.1 is stronger here---its GPQA
majority accuracy is slightly \emph{lower} than mini's ($\overline{\Mlab}{=}0.48$ vs.\ $0.52$)---but
because it is over-confident: mean $\C{=}0.89$, reaching $\C\geq0.8$ on $77\%$ of GPQA cases. With
agreement piled at the ceiling, $\C$ loses discriminative power. gpt-4.1 also has the worst GPQA
calibration of all twelve cells ($\mathrm{ECE}=0.41$, both fixed-width and equal-mass;
Brier $0.42$) versus $0.22$ for CoT-mini (Table~\ref{tab:cal}; Figures~\ref{fig:reliability},
\ref{fig:core}). The miscalibration is not GPQA-specific: on AIME gpt-4.1's ECE is $0.30$ at majority accuracy only $0.17$.

\begin{table}[t]
\centering\footnotesize
\setlength{\tabcolsep}{3pt}
\begin{tabular}{@{}lccccc@{}}
\toprule
GPQA condition & ECE & Brier & $\Pr[\C{\geq}.8]$ & err$\mid\C{\geq}.8$ & always-CW \\
\midrule
nano            & .36 & .37 & .50 & \textbf{.50} & .09 \\
mini (A)        & .33 & .35 & .67 & .43 & .15 \\
mini ZS (B)     & .30 & .33 & .59 & .39 & .11 \\
mini CoT (B)    & \textbf{.22} & \textbf{.26} & .59 & .28 & \textbf{.08} \\
mini (C)        & .30 & .32 & .62 & .36 & .14 \\
\textbf{gpt-4.1}& \textbf{.41} & \textbf{.42} & \textbf{.77} & .48 & \textbf{.28} \\
\bottomrule
\end{tabular}
\caption{GPQA calibration and confident-wrongness. ECE is fixed-width (equal-mass agrees within
$\pm.01$; Appendix); Brier is the mean squared error of $\C$ against $\Mlab$. ``err$\mid\C{\geq}.8$''---read as the conditional
majority-error rate given self-consistency $\C\geq0.8$ (i.e.\ $\geq80\%$ agreement)---is
the row-weighted majority-error rate ($\Mlab{=}0$) among these high-agreement entries; ``always-CW'' is the
(case-weighted) fraction of unique cases confidently wrong ($\C{\geq}.8$ \emph{and} $\A{\leq}.3$---a
stricter sample-level notion than majority-error) for \emph{every} runner. gpt-4.1 is worst
on ECE, Brier, $\Pr[\C{\geq}.8]$, and always-CW; its conditional high-confidence error ($.48$) is high
but marginally below nano's ($.50$). CoT-mini is best on ECE, Brier, and always-CW.}
\label{tab:cal}
\end{table}

The deployment consequence is concrete: \textbf{a high-agreement ($\C\geq0.8$) gpt-4.1 answer on GPQA
is wrong $48\%$ of the time} ($95\%$ CI $[.40,.56]$, case-clustered, row-weighted over high-agreement
entries), so a router that auto-trusts such answers inherits roughly that error rate. This direction is consistent with
a documented side effect of post-training \citep{openai2023gpt4}: alignment steps such as RLHF can make a
model concentrate its probability on a single answer---so it repeats that answer across samples, raising
$\C$---without making the answer more likely to be correct, sharpening apparent confidence faster than
accuracy; we do not identify the cause here. Our
contribution is not that this happens, but a quantification of how it degrades self-consistency
\emph{as a deployable signal}, and (\S\ref{sec:intrinsic}) that the resulting errors frequently recur
across different runners and prompts. These conclusions are not artifacts of the runner-clustered
bootstrap: under a \emph{case}-clustered bootstrap (resampling items) the frontier $\Delta\rho(\C,\Mlab)$
remains negative with a CI excluding zero (GPQA $-.15\,[-.27,-.03]$, AIME $-.21\,[-.32,-.12]$) and is
stable to leaving out any single runner.

\paragraph{A logged-snapshot confirmation.} Because the course runs lack recorded model snapshots
(\S\ref{sec:limitations}), we re-ran the Axis-C comparison ourselves under a \emph{single logged
snapshot} (\texttt{gpt-4.1-2025-04-14} vs.\ \texttt{gpt-4.1-mini-2025-04-14}, $K{=}20$, fixed canonical
prompt, the same $48$ GPQA cases). The pattern reproduces: gpt-4.1 is more self-consistent than mini
(mean $\C$ $0.89$ vs.\ $0.82$; $\C\geq0.8$ on $81\%$ vs.\ $63\%$) at \emph{identical} majority accuracy
($0.48$ each), with worse calibration ($\mathrm{ECE}=0.41$ vs.\ $0.34$) and lower $\rho(\C,\Mlab)$
($0.35$ vs.\ $0.41$; logged at $2026$-$05$-$24$T$11$:$53$Z). This makes a pure unlogged-snapshot explanation
less likely---though the re-run is small ($48$ GPQA, $K{=}20$) and the $\rho$/ECE \emph{gaps} are not
individually significant at this scale ($\Delta\rho{=}-.06\,[-.37,.26]$,
$\Delta\mathrm{ECE}{=}+.07\,[-.07,.22]$); the over-confidence itself (higher $\C$ at equal accuracy)
reproduces clearly.

\section{Chain-of-thought: accuracy yes, signal marginal}
\label{sec:cot}
Axis B isolates chain-of-thought on a fixed model (mini) with a within-runner paired design. CoT
robustly improves \emph{accuracy}: paired $\Delta\A=+0.067$ (GPQA, Wilcoxon signed-rank test---a paired significance test, not a correlation---$p{=}2.1{\times}10^{-5}$)
and $+0.069$ (AIME, $p{=}2.9{\times}10^{-4}$), with majority flips $2$--$5{\times}$ more often toward
correct than away. Its effect on the deployment label is weaker: $\Delta\Mlab=+0.066$ (GPQA; bootstrap
CI $[-0.005,0.132]$, Wilcoxon signed-rank $p{=}0.02$) and $+0.080$ (AIME, $p{=}0.003$). The effect on the
\emph{signal} is mixed: CoT raises $\rho(\C,\A)$ significantly on GPQA ($+0.199\,[0.049,0.367]$,
$p{=}0.0002$) but its effect on the primary $\rho(\C,\Mlab)$ is only borderline
($+0.123\,[-0.030,0.265]$, $p{=}0.025$ GPQA; $+0.031$, $p{=}0.21$ AIME). Under \emph{case}-clustered
resampling the GPQA $\Delta\rho(\C,\Mlab)$ CI excludes zero ($+.123\,[.016,.238]$) while AIME stays null
($+.031\,[-.023,.080]$). CoT is thus a reliable
accuracy intervention but at most a marginal one for the agreement--correctness signal.

\section{Self-consistency vs.\ other confidence signals}
\label{sec:signals}
We compare self-consistency against two signals from prior calibration work, on the same
gpt-4.1-mini cases ($50$ per benchmark). \emph{Verbalized confidence} \citep{lin2022teaching,tian2023justask}
is the mean of ten elicited $0$--$100$ self-ratings; \emph{P(True)} \citep{kadavath2022know} is the
True/False next-token logprob mass on the model's own modal answer. AURC (area under the
risk--coverage curve) and ECE target $\Mlab$; CIs are case-level (we ran these cases ourselves, so
runner clustering does not apply). Table~\ref{tab:signals} reports GPQA.

\begin{table}[t]
\centering\footnotesize
\begin{tabular}{@{}lcccc@{}}
\toprule
GPQA signal & cov. & $\rho(\cdot,\Mlab)\uparrow$ & AURC$\downarrow$ & ECE$\downarrow$ \\
\midrule
self-consistency $\C$ & 50/50 & \textbf{.31}\,[.05,.57] & .44 & .30 \\
verbalized conf.\ & 39/50 & .21\,[$-$.10,.49] & \textbf{.30}\textsuperscript{$\ast$} & .32 \\
P(True) & 50/50 & .19\,[$-$.10,.45] & .41 & .39 \\
rank-avg combo & 50/50 & .30\,[.03,.55] & .37 & \textbf{.13} \\
\bottomrule
\end{tabular}
\caption{Confidence signals on GPQA ($n{=}50$, gpt-4.1-mini). No signal dominates; differences sit
within wide CIs. Verbalized confidence parsed on only $39/50$ GPQA cases, so its $\rho$/AURC/ECE are
computed over those parsed cases and are \emph{not} directly comparable to $\C$/P(True), which use all
$50$ (``cov.''\ column); \textsuperscript{$\ast$}its AURC is computed on those $39$ cases, a favourable
subset, so the verbalized-confidence AURC advantage should be read with caution.}
\label{tab:signals}
\end{table}

\begin{figure}[t]
\centering
\includegraphics[width=\columnwidth]{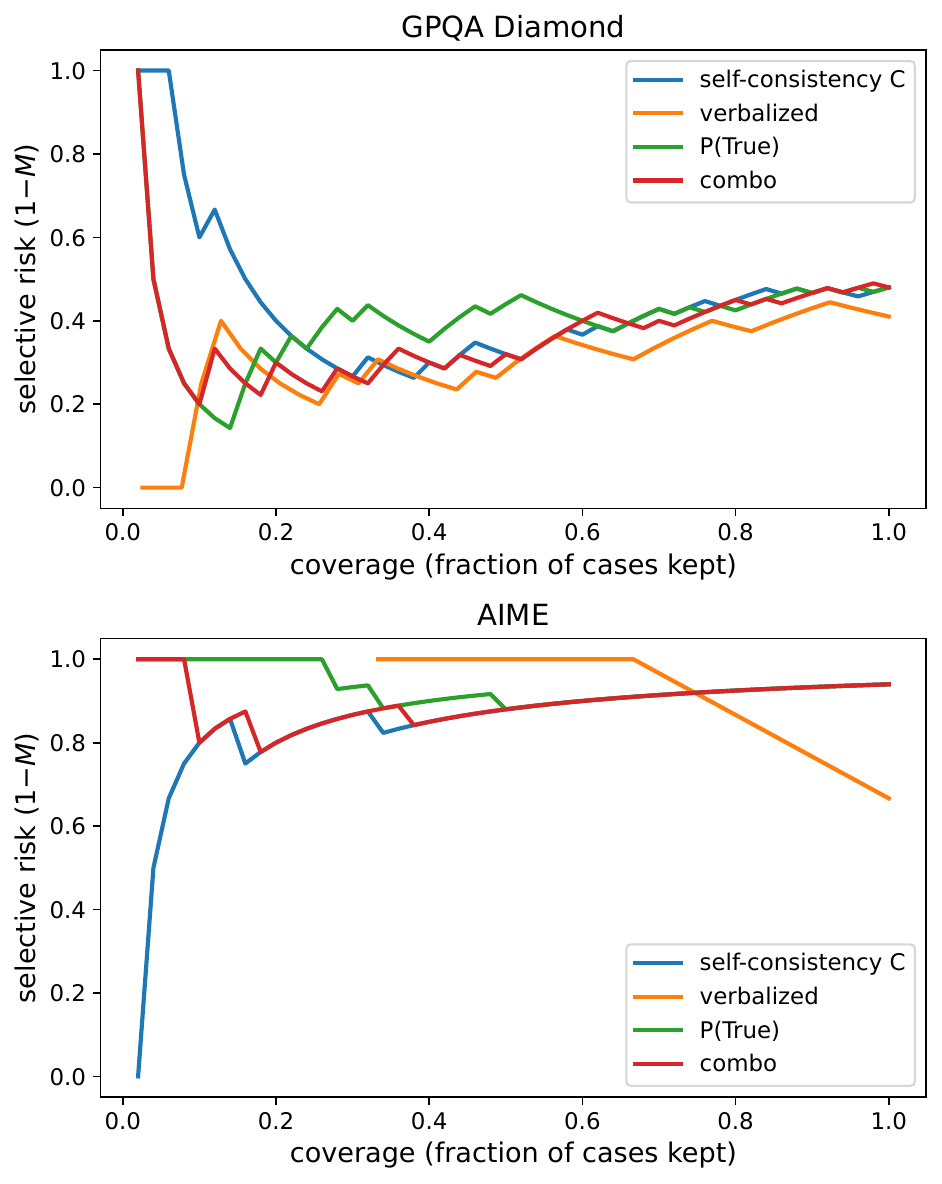}
\caption{Risk--coverage curves by confidence signal (gpt-4.1-mini). Lower is better. No signal is a
reliable standalone deferral score on these hard benchmarks.}
\label{fig:signals}
\end{figure}

On GPQA, self-consistency has the highest rank correlation with correctness ($\rho{=}0.31$), edging
verbalized confidence ($0.21$) and P(True) ($0.19$); yet verbalized confidence yields the best
risk--coverage (AURC $0.30$ vs.\ $0.44$) and a rank-average combination is best-calibrated (ECE
$0.13$), with all differences within wide CIs ($n{=}50$). On AIME every signal is weak---majority
accuracy is near the floor---and verbalized confidence is essentially uninformative: it parsed
on only $3/50$ AIME cases and was nearly constant (mean $0.92$), while P(True) is only weakly predictive
($\rho{=}0.12$). Self-consistency is thus \emph{competitive with, not dominated by}, dedicated
confidence signals on these hard benchmarks, and none is a reliable standalone abstention score
(Figure~\ref{fig:signals}).

\section{Confident errors recur---but are partly positional}
\label{sec:intrinsic}
Are confidently-wrong outcomes mere sampling artifacts? The replication design says not entirely:
several GPQA questions yield the same wrong majority answer at $\C{=}1.0$ from \emph{every} runner who
ran them, across different model conditions. For example, \texttt{gpqa\_d\_076} and
\texttt{gpqa\_d\_106} (both ground truth B) each elicited a wrong ``C'' from all runners across
multiple axes. Quantitatively, for gpt-4.1 on GPQA, $28\%$ of unique cases are confidently wrong for
\emph{every} runner and $50\%$ for \emph{at least one}---so a nontrivial subset of confident errors
recur across different runners and prompts rather than being sampling noise. We avoid the stronger
claim that they are wholly intrinsic: AIME confident errors are often dispersed (no single stable
wrong integer, frequently unparseable), and a minority of GPQA items appear ambiguous or mislabeled
(e.g.\ one item answered consistently against its listed key)---a caveat for treating every such case
as a model error.

\begin{figure}[t]
\centering
\includegraphics[width=\columnwidth]{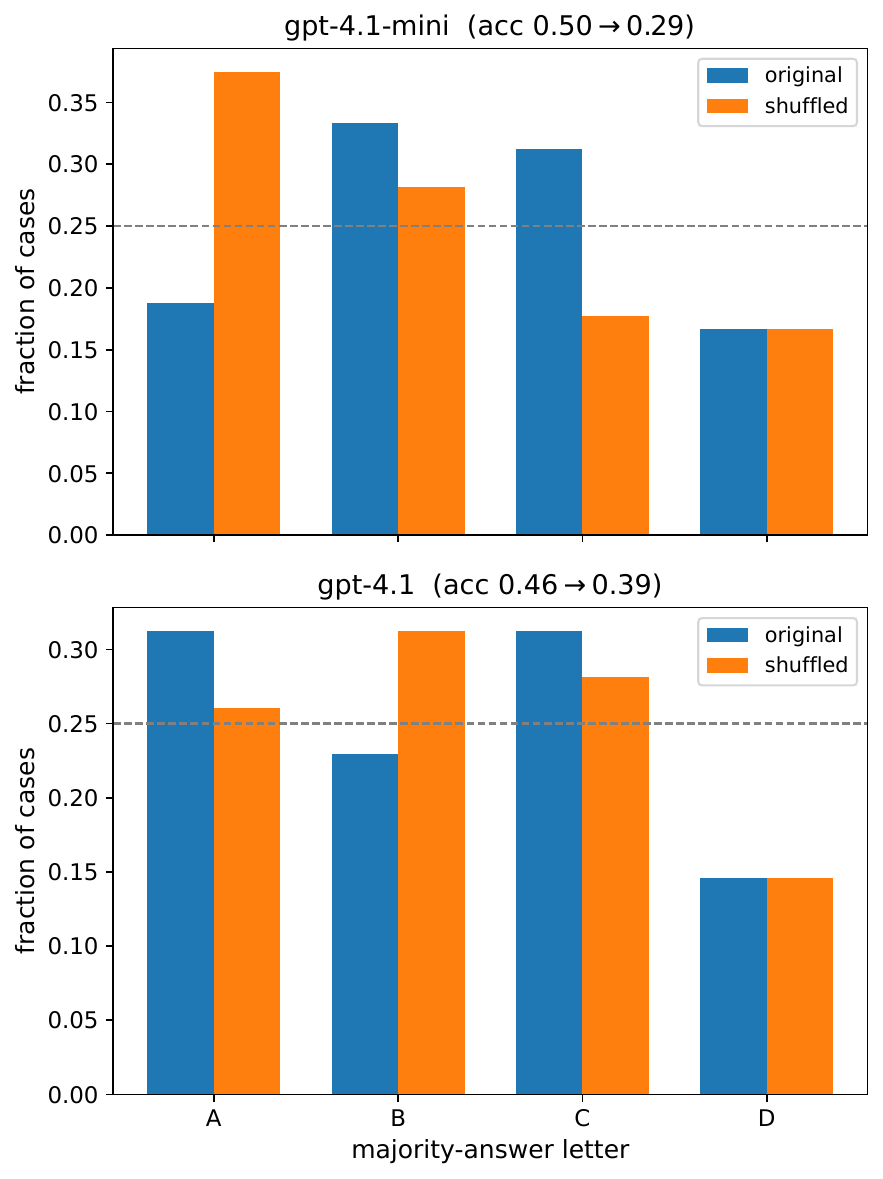}
\caption{Majority-answer letter distribution under original vs.\ shuffled options. The model
under-selects ``D'' regardless of which content sits there, and GPQA accuracy drops under shuffling.}
\label{fig:shuffle}
\end{figure}

\paragraph{An option-shuffle control.} Because GPQA is multiple choice, high agreement on a letter can
reflect position priors rather than semantic confidence \citep{zheng2024mcq,wei2024selection}. On $48$
GPQA cases ($12$ per ground-truth letter), for gpt-4.1-mini and gpt-4.1, we ran $K{=}50$ under the
original order and two random option permutations, remapping the ground-truth letter under each
(Figure~\ref{fig:shuffle}). Two effects emerge. First, answer content is \emph{position-sensitive}: mini's point-estimate
majority accuracy falls from $0.50\,[0.35,0.65]$ to $0.29\,[0.19,0.41]$ under shuffling (gpt-4.1 falls
less, $0.46{\to}0.39$)---suggestive, though the CIs overlap at $n{=}48$---and the majority \emph{content}
is stable across permutations for only $35\%$ of mini cases ($56\%$ for gpt-4.1). Second, ``D'' is
\emph{under-selected in both} the original and shuffled conditions ($\approx$$16\%$ for mini,
$\approx$$15\%$ for gpt-4.1, vs.\ a $25\%$ uniform baseline), so the D-avoidance is a position bias that
survives content randomization. A
meaningful share of GPQA ``confidence'' is therefore positional, not semantic.

\section{An exploratory cross-family check: are confident errors shared?}
\label{sec:crossfamily}
The audit so far is within one provider. As an \emph{exploratory} robustness check---\emph{not} a
like-for-like reproduction---we re-ran the GPQA and AIME measurements on three tiers of a second family,
Anthropic's Claude (\texttt{claude-haiku-4.5}, \texttt{claude-sonnet-4.6}, \texttt{claude-opus-4.7}), on the same studied cases ($48$ GPQA, $12$ per ground-truth
letter; $24$ AIME). The caveats are substantial and we state them first: the samples are drawn as separate, independent
queries, which we treat as approximately independent but cannot control for sampling temperature or verify
backend independence; $K{=}10$ per model is
far below the main study's $K{=}50$; and these models reason internally even when asked for a direct
answer (so AIME is effectively chain-of-thought---we therefore center the comparison on GPQA). Absolute
values are internal, not comparable to the gpt-4.1 runs; CIs are case-level bootstraps. With those
caveats, the qualitative pattern recurs.

\paragraph{Frontier over-confidence recurs.} Table~\ref{tab:crossfamily} shows the same
non-monotonic pattern within Claude. The frontier tier (opus) is the \emph{most} self-consistent
(mean $\C{=}0.94$, $\C\geq0.8$ on $88\%$ of GPQA cases) but \emph{not} the most accurate---its majority
accuracy ($0.58$) trails the mid tier (sonnet, $0.63$)---and it has the worst calibration
($\mathrm{ECE}=0.35$) and a lower $\rho(\C,\Mlab)$ ($0.44$) than sonnet ($0.74$). As in the gpt-4.1
family, the most-consistent tier discriminates correctness \emph{worse} than the mid tier.

\begin{table}[t]
\centering\footnotesize
\setlength{\tabcolsep}{3pt}
\begin{tabular}{@{}lccccc@{}}
\toprule
GPQA (Claude) & $\overline{\C}$ & $\overline{\Mlab}$ & $\rho(\C,\Mlab)$\,[CI] & ECE & err$\mid\C{\geq}.8$ \\
\midrule
haiku           & .80 & .50 & .29\,[.00,.57] & .34 & .43 \\
sonnet          & .87 & \textbf{.63} & \textbf{.74}\,[.54,.90] & \textbf{.24} & \textbf{.20} \\
\textbf{opus}   & \textbf{.94} & .58 & .44\,[.19,.67] & .35 & .36 \\
\bottomrule
\end{tabular}
\caption{Exploratory cross-family check on Claude models (GPQA, $K{=}10$ independent samples per
case, $48$ cases). The frontier tier (opus) is the most self-consistent and most over-confident
($\C\geq0.8$ on $88\%$) yet not the most accurate and the worst-calibrated---reproducing the gpt-4.1
pattern (Table~\ref{tab:cal}).}
\label{tab:crossfamily}
\end{table}

\paragraph{Confident errors are partly shared across families.} On the $46$ GPQA items the gpt-4.1 runs
also cover, gpt-4.1 and the Claude tiers agree on the majority answer for $48$--$63\%$ of items. Because
option-position priors (\S\ref{sec:intrinsic}) make a flat $1/3$ null inappropriate, we test the
same-wrong-answer rate (among items \emph{both} get wrong) against a label-permutation null that preserves
each model's empirical wrong-answer distribution (Table~\ref{tab:shared}). The observed rate exceeds the
null for all three tiers and significantly so for \texttt{claude-haiku-4.5} ($p{=}.003$) and
\texttt{claude-sonnet-4.6} ($p{=}.005$), though not \texttt{claude-opus-4.7} ($p{=}.07$, i.e.\ not
statistically significant at the conventional $p<.05$ threshold; only $12$ shared-wrong items); and these shared wrong answers are \emph{high-confidence}
for both providers (mean $\C\approx0.85$ GPT, $\approx0.78$ Claude), so they are confident errors, not
low-agreement coincidences. Six of the $46$ items are answered wrongly by \emph{all four} models (e.g.\
one whose listed key is B that every model answers D). Confident errors thus appear partly \emph{shared}
across providers rather than idiosyncratic---consistent with shared pretraining biases, shared
misconceptions, or item flaws---which both strengthens the recurrence finding of \S\ref{sec:intrinsic} and
cautions that cross-family agreement is itself not evidence of correctness (\S\ref{sec:limitations}).

\begin{table}[t]
\centering\footnotesize
\setlength{\tabcolsep}{4pt}
\begin{tabular}{@{}lccccc@{}}
\toprule
vs.\ gpt-4.1 & both-wrong & same & obs & null & $p$ \\
\midrule
haiku  & 15 & 10 & .67 & .31 & .003 \\
sonnet & 14 & 10 & .71 & .38 & .005 \\
opus   & 12 & \phantom{0}6 & .50 & .27 & .07\phantom{0} \\
\bottomrule
\end{tabular}
\caption{Shared confident errors (GPQA, $46$ shared items, exploratory). Among items \emph{both} gpt-4.1
and the Claude tier answer wrongly (``both-wrong''), how often they pick the \emph{same} wrong option
(``same''; obs $=$ same/both-wrong), vs.\ a label-permutation null preserving each model's wrong-answer
marginal ($p$ is the permutation tail). Shared-wrong items are high-confidence for both (mean $\C$
$0.84$--$0.93$ gpt-4.1, $0.70$--$0.87$ Claude).}
\label{tab:shared}
\end{table}

\section{Implications for systems}
\label{sec:systems}
\paragraph{Adaptive sampling.} Simulating stop-when-consensus on the existing sample banks, an adaptive
rule uses $60\%$ fewer samples on average ($19.8$ vs.\ $50$) at \emph{equal} majority error ($0.603$);
on CoT-mini the saving is $82\%$. Agreement is thus useful for \emph{allocating} compute even where it
is weak for \emph{trusting} an answer, consistent with \citet{aggarwal2023adaptive}.
\paragraph{Confidence-routed cascade.} Routing cheap$\to$expensive by $\C$ \citep{chen2023frugalgpt}
beats random ordering on the cost--accuracy frontier (peak majority accuracy $0.36$ vs.\ $0.34$) but is
\emph{dominated by simply always using the mid-tier model} ($0.40$); always-frontier reaches only
$0.33$, and an oracle router $0.44$. Because escalation sends ``hard'' (low-$\C$) cases to the
over-confident, no-more-accurate frontier model (\S\ref{sec:frontier}), routing toward it is
counterproductive here---a direct, cautionary consequence of the frontier degradation (Table~\ref{tab:cascade}).

\begin{table}[t]
\centering\footnotesize
\begin{tabular}{@{}lc@{}}
\toprule
Routing policy (GPQA, axes A+C) & majority accuracy \\
\midrule
always-nano & .30 \\
\textbf{always-mini} & \textbf{.40} \\
always-gpt-4.1 & .33 \\
$\C$-routed cascade & .30--.36 \\
random routing & .30--.34 \\
oracle routing & .44 \\
\bottomrule
\end{tabular}
\caption{Cost-aware routing. The $\C$-routed cascade beats random ordering but is dominated by simply
always using the mid-tier model; escalating to the over-confident frontier model hurts.}
\label{tab:cascade}
\end{table}

\section{Related work}
\label{sec:related}
Self-consistency was introduced to raise accuracy via majority voting \citep{wang2023selfconsistency},
not as a calibrated confidence; reusing the agreement signal as confidence is the downstream practice
we audit. LLM confidence can be elicited and is often miscalibrated---via self-evaluation/P(True)
\citep{kadavath2022know}, verbalized confidence \citep{lin2022teaching,tian2023justask}, and
post-training calibration regression \citep{openai2023gpt4}. Agreement already drives adaptive sampling
\citep{aggarwal2023adaptive} and confidence cascades \citep{chen2023frugalgpt}; selective prediction
\citep{geifman2017selective,kamath2020selective} thresholds a confidence signal to abstain.
Multiple-choice answers carry option-order/selection biases
\citep{zheng2024mcq,wei2024selection,pezeshkpour2024sensitivity}. Closest to our concern, sampling-based
agreement is itself used as an uncertainty or hallucination signal---SelfCheckGPT
\citep{manakul2023selfcheckgpt} and semantic-entropy methods \citep{kuhn2023semantic,farquhar2024semantic}---and
surveys of LLM confidence estimation treat consistency as one such cue \citep{geng2024survey,xiong2024canllms}.
A separate line shows that a model's preference for its \emph{own} outputs---self-enhancement/self-preference
bias in LLM-as-judge settings \citep{zheng2023judging,panickssery2024llmevaluators}---can inflate
self-agreement independently of correctness; self-consistency is the intra-model analogue, which our
cross-family shared-error analysis (\S\ref{sec:crossfamily}) probes directly. Judging correctness when
reference labels are scarce or contested is itself an active measurement problem
\citep{ding2025variance}, and our runner-clustered, heterogeneity-aware inference is in the spirit of
clustered estimation over heterogeneous units \citep{ding2024causal,ding2025causal}. Our contribution is an
empirical audit of the agreement signal in the confidence role on hard reasoning benchmarks, with a
cross-runner recurrence analysis and an exploratory cross-family check that separates recurring from
stochastic confident-wrongness and finds confident errors are partly shared across providers.

\section{Conclusion}
\label{sec:conclusion}
Self-consistency is \emph{not} accuracy, but it is a positive, regime-dependent empirical proxy for it:
useful where agreement is unsaturated---mid-tier models, and for allocating sample budget---but unreliable
as a standalone confidence score. The agreement--correctness correlation is positive but weak throughout
(Spearman $\rho$ $0.20$--$0.59$), and weakest for the over-confident frontier model, whose high-agreement
($\C\geq0.8$) GPQA answers are still wrong $48\%$ of the time. Along the way it is only marginally improved by chain-of-thought on the
deployment label, and partly positional on multiple choice; and its confident errors frequently recur
across runners and prompts---and, in an exploratory cross-family check, across providers---so they reflect
shared bias more than sampling noise. The practical upshot is not ``never use agreement'' but ``use it
within its regime'': treat it as a coarse, cost-saving signal, not as a trustworthy confidence threshold,
and never as a reason to route hard cases toward the most consistent frontier model.

\section*{Limitations}
\label{sec:limitations}
Our primary audit is within a single provider (gpt-4.1); we add an exploratory cross-family check on
Claude models (\S\ref{sec:crossfamily}), but it uses a separate sampling path---no \texttt{temperature} control,
smaller $K$, and internal reasoning we cannot disable---so its absolute values are internal, not
directly comparable to the gpt-4.1 runs. Crucially, because the confident errors are largely
\emph{shared} across families, cross-family agreement does not by itself certify correctness: shared
pretraining data and shared option-position priors can produce shared bias rather than independent
confirmation. We study two benchmarks (four-option multiple-choice questions and integer math), not open-ended generation or code.
$K{=}50$ leaves substantial measurement noise. We do not study reasoning-trained models, whose internal
reasoning may already supply the calibration lift we measure for prompted CoT. Runners wrote their own
prompts, adding implementation variance. Critically, exact model snapshots and run timestamps were not
logged and are unrecoverable, so we cannot verify that the replicated runs occurred under an identical
backend snapshot; our design establishes recurrence across \emph{different} runners and prompts, not
across time---though a controlled re-run under a single logged snapshot (\S\ref{sec:frontier},
$2026$-$05$-$24$) reproduces the over-confidence direction (higher agreement at equal accuracy), while
the $\rho$/ECE gaps are underpowered at $K{=}20$. Finally, the confidence-baseline and shuffle studies use $50$ and $48$ cases respectively, so their CIs are wide,
and a minority of GPQA items appear ambiguous or mislabeled, which inflates apparent confident-wrongness.
The confidence-baseline and option-shuffle controls were run by us under a current model snapshot, so
their absolute values are internal comparisons, not directly comparable to the course runs.

\section*{Acknowledgments}
I thank the graduate-course cohort whose submitted model-output runs constitute the dataset analyzed
here. All analysis, framing, and conclusions are the author's own.

The author acknowledges the National Artificial Intelligence Research Resource (NAIRR) Pilot and the
Jetstream2 cloud resource at Indiana University for contributing to this research result. This work
used Jetstream2 through NAIRR Pilot allocation NAIRR250223; Jetstream2 is supported by the U.S.\
National Science Foundation under award NSF-OAC~2005506. The author also thanks the Wharton AI \&
Analytics Initiative at the University of Pennsylvania for financially supporting this project through
its AI Education Innovation Fund.

\bibliography{refs}

@inproceedings{wang2023selfconsistency,
  title     = {Self-Consistency Improves Chain of Thought Reasoning in Language Models},
  author    = {Wang, Xuezhi and Wei, Jason and Schuurmans, Dale and Le, Quoc V. and Chi, Ed H. and Narang, Sharan and Chowdhery, Aakanksha and Zhou, Denny},
  booktitle = {The Eleventh International Conference on Learning Representations (ICLR)},
  year      = {2023},
  url       = {https://openreview.net/forum?id=1PL1NIMMrw},
  eprint    = {2203.11171},
  archivePrefix = {arXiv},
  primaryClass  = {cs.CL},
  note      = {arXiv:2203.11171}
}

@inproceedings{wei2022cot,
  title     = {Chain-of-Thought Prompting Elicits Reasoning in Large Language Models},
  author    = {Wei, Jason and Wang, Xuezhi and Schuurmans, Dale and Bosma, Maarten and Ichter, Brian and Xia, Fei and Chi, Ed H. and Le, Quoc V. and Zhou, Denny},
  booktitle = {Advances in Neural Information Processing Systems 35 (NeurIPS)},
  year      = {2022},
  url       = {https://proceedings.neurips.cc/paper_files/paper/2022/hash/9d5609613524ecf4f15af0f7b31abca4-Abstract-Conference.html},
  eprint    = {2201.11903},
  archivePrefix = {arXiv},
  primaryClass  = {cs.CL},
  note      = {arXiv:2201.11903}
}

@inproceedings{rein2024gpqa,
  title     = {{GPQA}: A Graduate-Level Google-Proof {Q\&A} Benchmark},
  author    = {Rein, David and Hou, Betty Li and Stickland, Asa Cooper and Petty, Jackson and Pang, Richard Yuanzhe and Dirani, Julien and Michael, Julian and Bowman, Samuel R.},
  booktitle = {First Conference on Language Modeling (COLM)},
  year      = {2024},
  url       = {https://openreview.net/forum?id=Ti67584b98},
  eprint    = {2311.12022},
  archivePrefix = {arXiv},
  primaryClass  = {cs.AI},
  note      = {arXiv:2311.12022}
}

@misc{aime,
  title        = {The {American Invitational Mathematics Examination} ({AIME})},
  author       = {{MAA}},
  year         = {2024},
  howpublished = {Mathematical Association of America},
  note         = {\url{https://maa.org}}
}

@article{kadavath2022know,
  title   = {Language Models (Mostly) Know What They Know},
  author  = {Kadavath, Saurav and Conerly, Tom and Askell, Amanda and Henighan, Tom and Drain, Dawn and Perez, Ethan and Schiefer, Nicholas and Hatfield-Dodds, Zac and DasSarma, Nova and Tran-Johnson, Eli and Johnston, Scott and El-Showk, Sheer and Jones, Andy and Elhage, Nelson and Hume, Tristan and Chen, Anna and Bai, Yuntao and Bowman, Sam and Fort, Stanislav and Ganguli, Deep and Hernandez, Danny and Jacobson, Josh and Kernion, Jackson and Kravec, Shauna and Lovitt, Liane and Ndousse, Kamal and Olsson, Catherine and Ringer, Sam and Amodei, Dario and Brown, Tom and Clark, Jack and Joseph, Nicholas and Mann, Ben and McCandlish, Sam and Olah, Chris and Kaplan, Jared},
  journal = {arXiv preprint arXiv:2207.05221},
  year    = {2022},
  url     = {https://arxiv.org/abs/2207.05221},
  eprint  = {2207.05221},
  archivePrefix = {arXiv},
  primaryClass  = {cs.CL}
}

@inproceedings{tian2023justask,
  title     = {Just Ask for Calibration: Strategies for Eliciting Calibrated Confidence Scores from Language Models Fine-Tuned with Human Feedback},
  author    = {Tian, Katherine and Mitchell, Eric and Zhou, Allan and Sharma, Archit and Rafailov, Rafael and Yao, Huaxiu and Finn, Chelsea and Manning, Christopher D.},
  booktitle = {Proceedings of the 2023 Conference on Empirical Methods in Natural Language Processing (EMNLP)},
  pages     = {5433--5442},
  year      = {2023},
  publisher = {Association for Computational Linguistics},
  address   = {Singapore},
  url       = {https://aclanthology.org/2023.emnlp-main.330/},
  doi       = {10.18653/v1/2023.emnlp-main.330},
  eprint    = {2305.14975},
  archivePrefix = {arXiv},
  primaryClass  = {cs.CL}
}

@article{lin2022teaching,
  title   = {Teaching Models to Express Their Uncertainty in Words},
  author  = {Lin, Stephanie and Hilton, Jacob and Evans, Owain},
  journal = {Transactions on Machine Learning Research (TMLR)},
  year    = {2022},
  url     = {https://openreview.net/forum?id=8s8K2UZGTZ},
  eprint  = {2205.14334},
  archivePrefix = {arXiv},
  primaryClass  = {cs.CL},
  note    = {arXiv:2205.14334}
}

@article{openai2023gpt4,
  title   = {{GPT-4} Technical Report},
  author  = {{OpenAI}},
  journal = {arXiv preprint arXiv:2303.08774},
  year    = {2023},
  url     = {https://arxiv.org/abs/2303.08774},
  eprint  = {2303.08774},
  archivePrefix = {arXiv},
  primaryClass  = {cs.CL}
}

@inproceedings{aggarwal2023adaptive,
  title     = {Let's Sample Step by Step: Adaptive-Consistency for Efficient Reasoning and Coding with {LLM}s},
  author    = {Aggarwal, Pranjal and Madaan, Aman and Yang, Yiming and Mausam},
  booktitle = {Proceedings of the 2023 Conference on Empirical Methods in Natural Language Processing (EMNLP)},
  pages     = {12375--12396},
  year      = {2023},
  publisher = {Association for Computational Linguistics},
  address   = {Singapore},
  url       = {https://aclanthology.org/2023.emnlp-main.761/},
  doi       = {10.18653/v1/2023.emnlp-main.761},
  eprint    = {2305.11860},
  archivePrefix = {arXiv},
  primaryClass  = {cs.CL}
}

@article{chen2023frugalgpt,
  title   = {{FrugalGPT}: How to Use Large Language Models While Reducing Cost and Improving Performance},
  author  = {Chen, Lingjiao and Zaharia, Matei and Zou, James},
  journal = {arXiv preprint arXiv:2305.05176},
  year    = {2023},
  url     = {https://arxiv.org/abs/2305.05176},
  eprint  = {2305.05176},
  archivePrefix = {arXiv},
  primaryClass  = {cs.LG}
}

@inproceedings{geifman2017selective,
  title     = {Selective Classification for Deep Neural Networks},
  author    = {Geifman, Yonatan and El-Yaniv, Ran},
  booktitle = {Advances in Neural Information Processing Systems 30 (NIPS)},
  year      = {2017},
  url       = {https://proceedings.neurips.cc/paper_files/paper/2017/hash/4a8423d5e91fda00bb7e46540e2b0cf1-Abstract.html},
  eprint    = {1705.08500},
  archivePrefix = {arXiv},
  primaryClass  = {cs.LG},
  note      = {arXiv:1705.08500}
}

@inproceedings{kamath2020selective,
  title     = {Selective Question Answering under Domain Shift},
  author    = {Kamath, Amita and Jia, Robin and Liang, Percy},
  booktitle = {Proceedings of the 58th Annual Meeting of the Association for Computational Linguistics (ACL)},
  pages     = {5684--5696},
  year      = {2020},
  publisher = {Association for Computational Linguistics},
  url       = {https://aclanthology.org/2020.acl-main.503/},
  doi       = {10.18653/v1/2020.acl-main.503},
  eprint    = {2006.09462},
  archivePrefix = {arXiv},
  primaryClass  = {cs.CL}
}

@inproceedings{zheng2024mcq,
  title     = {Large Language Models Are Not Robust Multiple Choice Selectors},
  author    = {Zheng, Chujie and Zhou, Hao and Meng, Fandong and Zhou, Jie and Huang, Minlie},
  booktitle = {The Twelfth International Conference on Learning Representations (ICLR)},
  year      = {2024},
  url       = {https://openreview.net/forum?id=shr9PXz7T0},
  eprint    = {2309.03882},
  archivePrefix = {arXiv},
  primaryClass  = {cs.CL},
  note      = {arXiv:2309.03882}
}

@inproceedings{wei2024selection,
  title     = {Unveiling Selection Biases: Exploring Order and Token Sensitivity in Large Language Models},
  author    = {Wei, Sheng-Lun and Wu, Cheng-Kuang and Huang, Hen-Hsen and Chen, Hsin-Hsi},
  booktitle = {Findings of the Association for Computational Linguistics: ACL 2024},
  pages     = {5598--5621},
  year      = {2024},
  publisher = {Association for Computational Linguistics},
  url       = {https://aclanthology.org/2024.findings-acl.333/},
  eprint    = {2406.03009},
  archivePrefix = {arXiv},
  primaryClass  = {cs.CL}
}

@inproceedings{pezeshkpour2024sensitivity,
  title     = {Large Language Models Sensitivity to The Order of Options in Multiple-Choice Questions},
  author    = {Pezeshkpour, Pouya and Hruschka, Estevam},
  booktitle = {Findings of the Association for Computational Linguistics: NAACL 2024},
  year      = {2024},
  publisher = {Association for Computational Linguistics},
  url       = {https://aclanthology.org/2024.findings-naacl.130/},
  eprint    = {2308.11483},
  archivePrefix = {arXiv},
  primaryClass  = {cs.CL}
}

@inproceedings{manakul2023selfcheckgpt,
  title     = {{SelfCheckGPT}: Zero-Resource Black-Box Hallucination Detection for Generative Large Language Models},
  author    = {Manakul, Potsawee and Liusie, Adian and Gales, Mark J. F.},
  booktitle = {Proceedings of the 2023 Conference on Empirical Methods in Natural Language Processing (EMNLP)},
  pages     = {9004--9017},
  month     = dec,
  year      = {2023},
  address   = {Singapore},
  publisher = {Association for Computational Linguistics},
  url       = {https://aclanthology.org/2023.emnlp-main.557/},
  eprint    = {2303.08896},
  archivePrefix = {arXiv},
  primaryClass  = {cs.CL}
}

@inproceedings{kuhn2023semantic,
  title     = {Semantic Uncertainty: Linguistic Invariances for Uncertainty Estimation in Natural Language Generation},
  author    = {Kuhn, Lorenz and Gal, Yarin and Farquhar, Sebastian},
  booktitle = {The Eleventh International Conference on Learning Representations (ICLR)},
  year      = {2023},
  url       = {https://openreview.net/forum?id=VD-AYtP0dve},
  eprint    = {2302.09664},
  archivePrefix = {arXiv},
  primaryClass  = {cs.CL}
}

@article{farquhar2024semantic,
  title     = {Detecting hallucinations in large language models using semantic entropy},
  author    = {Farquhar, Sebastian and Kossen, Jannik and Kuhn, Lorenz and Gal, Yarin},
  journal   = {Nature},
  volume    = {630},
  number    = {8017},
  pages     = {625--630},
  year      = {2024},
  publisher = {Nature Publishing Group},
  doi       = {10.1038/s41586-024-07421-0},
  url       = {https://www.nature.com/articles/s41586-024-07421-0}
}

@inproceedings{zheng2023judging,
  title     = {Judging {LLM}-as-a-Judge with {MT-Bench} and Chatbot Arena},
  author    = {Zheng, Lianmin and Chiang, Wei-Lin and Sheng, Ying and Zhuang, Siyuan and Wu, Zhanghao and Zhuang, Yonghao and Lin, Zi and Li, Zhuohan and Li, Dacheng and Xing, Eric P. and Zhang, Hao and Gonzalez, Joseph E. and Stoica, Ion},
  booktitle = {Advances in Neural Information Processing Systems 36 (NeurIPS 2023), Datasets and Benchmarks Track},
  year      = {2023},
  url       = {https://papers.nips.cc/paper_files/paper/2023/hash/91f18a1287b398d378ef22505bf41832-Abstract-Datasets_and_Benchmarks.html},
  eprint    = {2306.05685},
  archivePrefix = {arXiv},
  primaryClass  = {cs.CL}
}

@inproceedings{panickssery2024llmevaluators,
  title     = {{LLM} Evaluators Recognize and Favor Their Own Generations},
  author    = {Panickssery, Arjun and Bowman, Samuel R. and Feng, Shi},
  booktitle = {Advances in Neural Information Processing Systems 37 (NeurIPS 2024)},
  year      = {2024},
  url       = {https://proceedings.neurips.cc/paper_files/paper/2024/hash/7f1f0218e45f5414c79c0679633e47bc-Abstract-Conference.html},
  eprint    = {2404.13076},
  archivePrefix = {arXiv},
  primaryClass  = {cs.CL}
}

@inproceedings{geng2024survey,
  title     = {A Survey of Confidence Estimation and Calibration in Large Language Models},
  author    = {Geng, Jiahui and Cai, Fengyu and Wang, Yuxia and Koeppl, Heinz and Nakov, Preslav and Gurevych, Iryna},
  booktitle = {Proceedings of the 2024 Conference of the North American Chapter of the Association for Computational Linguistics: Human Language Technologies (NAACL)},
  pages     = {6577--6595},
  month     = jun,
  year      = {2024},
  address   = {Mexico City, Mexico},
  publisher = {Association for Computational Linguistics},
  url       = {https://aclanthology.org/2024.naacl-long.366/},
  eprint    = {2311.08298},
  archivePrefix = {arXiv},
  primaryClass  = {cs.CL}
}

@inproceedings{xiong2024canllms,
  title     = {Can {LLMs} Express Their Uncertainty? An Empirical Evaluation of Confidence Elicitation in {LLMs}},
  author    = {Xiong, Miao and Hu, Zhiyuan and Lu, Xinyang and Li, Yifei and Fu, Jie and He, Junxian and Hooi, Bryan},
  booktitle = {The Twelfth International Conference on Learning Representations (ICLR)},
  year      = {2024},
  url       = {https://openreview.net/forum?id=gjeQKFxFpZ},
  eprint    = {2306.13063},
  archivePrefix = {arXiv},
  primaryClass  = {cs.CL}
}

@phdthesis{ding2018dissertation,
  author = {Ding, Kaihua},
  title  = {{Efficient Output-Based Adaptation Mechanics for High-Order Computational Fluid Dynamics Methods}},
  school = {University of Michigan},
  year   = {2018},
  url    = {https://deepblue.lib.umich.edu/handle/2027.42/144065}
}

@misc{ding2025variance,
  author        = {Ding, Kaihua},
  title         = {{Variance-Bounded Evaluation of Entity-Centric AI Systems Without Ground Truth: Theory and Measurement}},
  year          = {2025},
  eprint        = {2509.22751},
  archivePrefix = {arXiv},
  primaryClass  = {stat.ML},
  doi           = {10.48550/arXiv.2509.22751}
}

@article{ding2025causal,
  author  = {Ding, Kaihua and Cui, Jingsong and Soltani, Mohammad and Jin, Jing},
  title   = {{Iterative Causal Segmentation}},
  journal = {PMSA Journal},
  pages   = {21--33},
  year    = {2025}
}

@misc{ding2024causal,
  author        = {Ding, Kaihua and Cui, Jingsong and Soltani, Mohammad and Jin, Jing},
  title         = {{Iterative Causal Segmentation: Filling the Gap between Market Segmentation and Marketing Strategy}},
  year          = {2024},
  eprint        = {2405.14743},
  archivePrefix = {arXiv},
  primaryClass  = {cs.LG},
  doi           = {10.48550/arXiv.2405.14743}
}

@misc{ding2025airesilient,
  author        = {Ding, Kaihua},
  title         = {{Designing AI-Resilient Assessments Using Interconnected Problems: A Theoretically Grounded and Empirically Validated Framework}},
  year          = {2025},
  eprint        = {2512.10758},
  archivePrefix = {arXiv},
  primaryClass  = {cs.AI},
  note          = {Accepted, IEEE Frontiers in Education (FIE) 2026},
  doi           = {10.48550/arXiv.2512.10758}
}

@misc{verga2024poll,
  author        = {Verga, Pat and Hofst\"atter, Sebastian and Althammer, Sophia and Su, Yixuan and Piktus, Aleksandra and Arkhangorodsky, Arkady and Xu, Minjie and White, Naomi and Lewis, Patrick},
  title         = {Replacing Judges with Juries: Evaluating {LLM} Generations with a Panel of Diverse Models},
  year          = {2024},
  eprint        = {2404.18796},
  archivePrefix = {arXiv},
  primaryClass  = {cs.CL}
}

@inproceedings{shazeer2017moe,
  author    = {Shazeer, Noam and Mirhoseini, Azalia and Maziarz, Krzysztof and Davis, Andy and Le, Quoc and Hinton, Geoffrey and Dean, Jeff},
  title     = {Outrageously Large Neural Networks: The Sparsely-Gated Mixture-of-Experts Layer},
  booktitle = {International Conference on Learning Representations (ICLR)},
  year      = {2017}
}

\appendix
\section{Hierarchical cluster bootstrap}
\label{app:boot}
For $B{=}2000$ replicates we resample the $53$ runner indices with replacement; within each sampled
runner we resample its cases with replacement; we recompute the statistic on the pooled resample and
take percentile CIs. This two-stage scheme is a hierarchical (nested) cluster bootstrap on the runner
factor: it corrects the dominant runner-level dependence rather than implementing a full two-way crossed
resample over globally shared cases. For the paired Axis-B contrasts we resample (runner, case) units while preserving
the zero-shot/CoT pairing and recompute the paired difference. Holm correction is applied across the
family of twelve per-cell $\rho(\C,\Mlab)$ tests; correlation-difference $p$-values are permutation-based.

\section{Prompts and protocol}
\label{app:prompts}
Canonical zero-shot and chain-of-thought prompts for GPQA and AIME, answer-extraction rules, and the
\texttt{\_UNPARSEABLE\_} handling (which counts as incorrect) are documented with the released code.
For the confidence baselines, verbalized confidence used a fixed ``Answer/Confidence'' template
(ten samples), and P(True) read the True/False token logprobs on the modal answer; for the shuffle
control, option contents were permuted with the ground-truth letter remapped accordingly.

\section{Additional tables and robustness}
\label{app:extra}
Per-cell marginal rates, reliability diagrams, full risk--coverage curves, equal-mass vs.\ fixed-width
ECE, the confident-wrongness threshold sweep, and the adaptive-sampling and cascade frontiers are
provided in the supplementary material.

\end{document}